\documentclass[]{oppo}

\usepackage[toc,page,header]{appendix}
\usepackage{fancyvrb}
\usepackage{wrapfig}
\usepackage{graphicx}
\usepackage{enumitem}
\usepackage{adjustbox}
\usepackage{pdfpages} 
\usepackage{amsmath} 
\usepackage{ragged2e}  
\usepackage{caption}
\newcommand{\method}{O-Mem}
\usepackage{hyperref}
\usepackage{url}
\usepackage{booktabs}
\usepackage{makecell} 
\usepackage{multirow}
\usepackage{xcolor}
\usepackage{xcolor} 
\usepackage{float}
\usepackage{listings}
\usepackage{fancyvrb}
\usepackage{tablefootnote}
\usepackage{wrapfig}
\usepackage{amsmath}  
\usepackage{amssymb} 
\usepackage{threeparttable}
\usepackage{booktabs}      
\usepackage{multirow}      
\usepackage{makecell}      
\usepackage{xcolor}        
\usepackage{colortbl}      
\usepackage{threeparttable} 
\usepackage[table, dvipsnames, svgnames, x11names ]{xcolor}
\usepackage{times}
\usepackage{latexsym}
\usepackage[T1]{fontenc}
\usepackage[utf8]{inputenc}
\usepackage{microtype}
\usepackage{inconsolata}
\usepackage{graphicx}
\usepackage{hyperref}       
\usepackage{url}            
\usepackage{booktabs}       
\usepackage{amsfonts}       
\usepackage{nicefrac}       
\usepackage{stackengine}
\usepackage{microtype}      
\usepackage{colortbl}
\usepackage{xcolor}
\usepackage{amsmath}
\usepackage{amssymb}
\usepackage{amsthm}
\usepackage{mathrsfs}
\usepackage{pifont}
\usepackage{MnSymbol}
\usepackage{balance}
\usepackage{enumitem}
\usepackage{xcolor}
\usepackage{natbib}
\usepackage{multicol}
\usepackage{xspace}
\usepackage{fancyvrb}
\usepackage{geometry}
\usepackage{tocloft}
\usepackage{tasks}
\usepackage{hyperref}
\usepackage{fontawesome5}

\AtBeginDocument{%
  \providecommand\BibTeX{{%
    \normalfont B\kern-0.5em{\scshape i\kern-0.25em b}\kern-0.8em\TeX}}}

\makeatletter
\DeclareRobustCommand\onedot{\futurelet\@let@token\@onedot}
\def\@onedot{\ifx\@let@token.\else.\null\fi}

\usepackage{setspace}
\usepackage{mathtools}

\usepackage{multirow,booktabs}
\usepackage{subcaption}

\usepackage[most]{tcolorbox}
\usepackage{amssymb}
\usepackage{color}
\usepackage{hyperref}  
\usepackage{graphicx}  
\definecolor{brightube}{rgb}{0.82, 0.62, 0.91}
\DeclareMathOperator*{\argmin}{arg\,min}  

\title{\method{}: Omni Memory System for Personalized, Long Horizon, Self-Evolving Agents}

\affiliation{OPPO AI Agent Team}

\abstract{
Recent advancements in LLM-powered agents have demonstrated significant potential in generating human-like responses; however, they continue to face challenges in maintaining long-term interactions within complex environments, primarily due to limitations in contextual consistency and dynamic personalization. Existing memory systems often depend on semantic grouping prior to retrieval, which can overlook semantically irrelevant yet critical user information and introduce retrieval noise. In this report, we propose the initial design of \method{}, a novel memory framework based on active user profiling that dynamically extracts and updates user characteristics and event records from their proactive interactions with agents. \method{} supports hierarchical retrieval of persona attributes and topic-related context, enabling more adaptive and coherent personalized responses. \method{} achieves 51.67\% on the public LoCoMo benchmark, a nearly 3\% improvement upon LangMem—the previous state-of-the-art, and it achieves 62.99\% on PERSONAMEM, a 3.5\% improvement upon A-Mem—the previous state-of-the-art. \method{} also boosts token and interaction response time efficiency compared to previous memory frameworks. Our work opens up promising directions for developing efficient and human-like personalized AI assistants in the future.}
\date{\today}

\correspondence{Wangchunshu Zhou at \email{zhouwangchunshu@oppo.com}}

\checkdata[Open Source]{%
  \href{https://github.com/OPPO-PersonalAI/O-Mem}{%
    \faGithub \ Github%
  }
\href{https://huggingface.co/papers/2511.13593}{%
    \faGithub \ Huggingface%
  }

}

\begin{document}
\maketitle

\begin{figure}[htbp]
  \centering
  \includegraphics[width=0.9\textwidth]{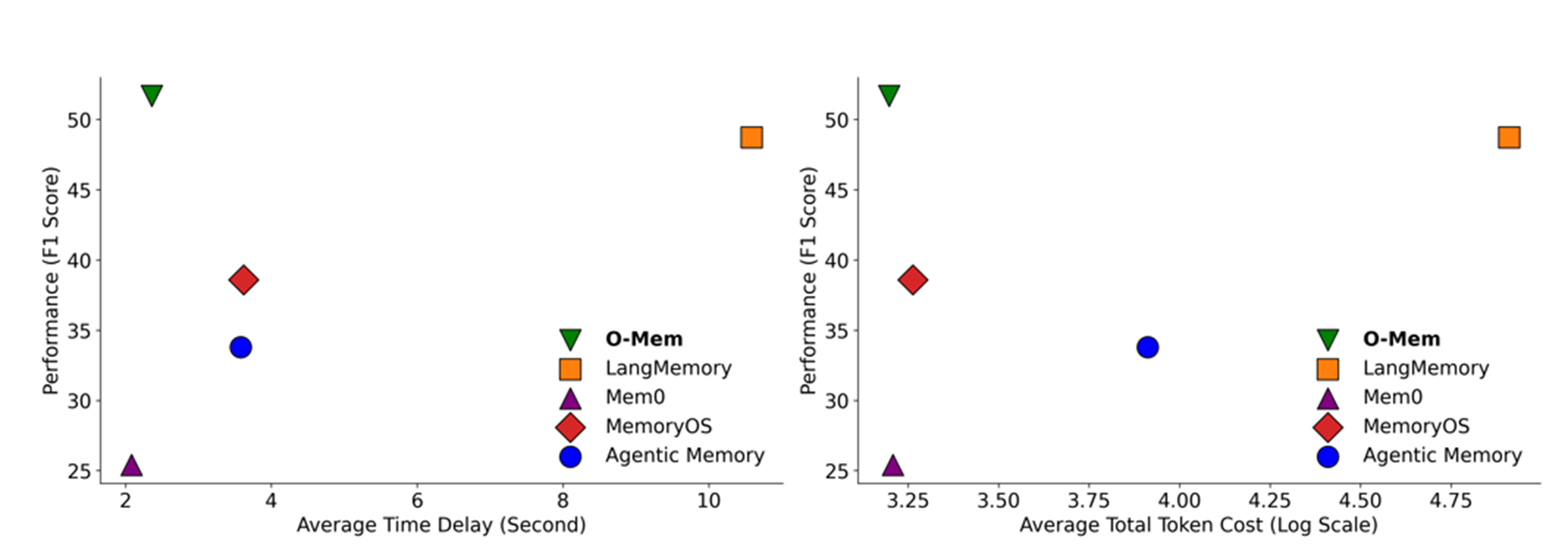}
  \caption{Trade-off between performance and efficiency of different memory frameworks. 
  (a) Left panel: Average latency per interaction (MemoryOS latency uses FAISS-CPU for compatibility, a conservative estimate). 
  (b) Right panel: Average computational cost (tokens) per interaction. 
  Results show \method{} achieves Pareto optimality in efficiency and performance. 
  Note: Token-control experiments were only conducted on LoCoMo's GPT-4.1; no token control for GPT-4o-mini and other two datasets.}
  \label{fig:compare_main}
\end{figure}


\clearpage
\setcounter{tocdepth}{3}

\section{Introduction}

LLM-empowered agents have demonstrated huge potential in generating human-level intelligent responses \citep{schlegel2025large} but still lack long-term interaction ability with complex external environments \citep{zhang2024survey}. This limitation causes agents to struggle to maintain consistency of context across time \citep{laban2025llms} and reduced their personalization capability dynamically adapting to users' situations \citep{zhang2025personaagent}.


Agent memory systems equip agents with the ability to retain and utilize past experiences, unlike conventional agents that treat each interaction as independent. These systems store user past interactions in diverse architectures and enable agents to retrieve relevant information from them to deliver more personalized responses. For example, Memory OS~\citep{kang2025memory} categorizes user interactions into short-term, mid-term, and long-term persona memory caches based on timestamps and frequency of occurrence. Agentic Memory~\citep{xu2025mem} organizes interactions into distinct groups according to their semantic similarity, while Mem0~\citep{chhikara2025mem0} extracts meaningful content from messages and stores the extracted information independently to support future retrieval. By structuring user information more effectively, these systems enhance the ability of agents to provide efficient and highly personalized responses.



\begin{figure}[t]
    \centering
    \includegraphics[width=1\textwidth]{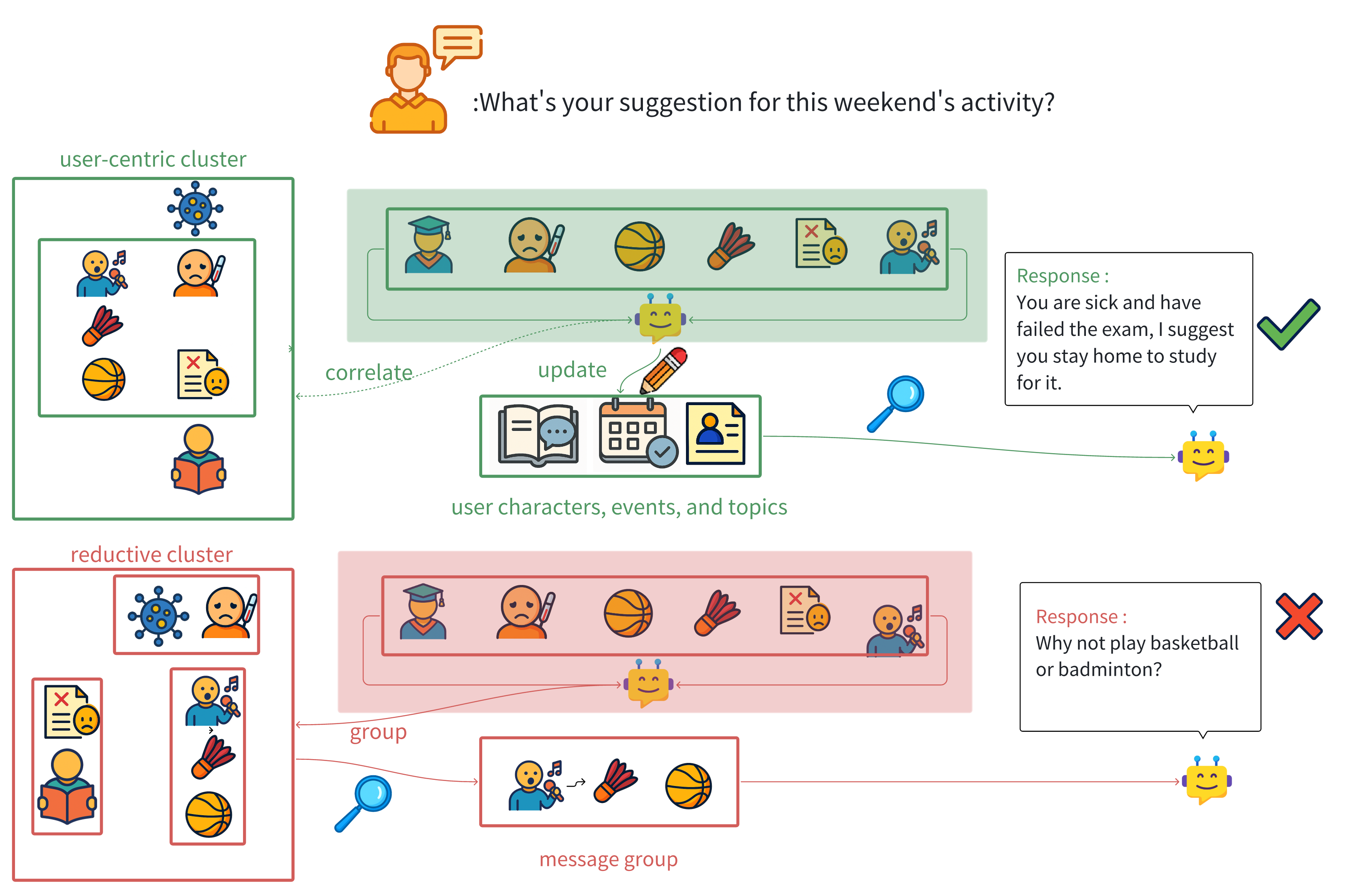}
    \caption{Top: our proposed user-centric framework O-Mem employing characteristic identification, event recording, and topic-message indexing. Bottom: the conventional memory system with semantic retrieval from message groupings. \method{} correlates virtual relationships among user's interactions (in dotted line).}
    \label{idea}
\end{figure}

The core pipeline of such memory systems involves grouping the messages based on semantic topics and retrieving memory groupings when interacting with users. However, this design presents several significant shortcomings: i) Memory systems that rely heavily on semantic retrieval may overlook information that is semantically irrelevant but potentially important—such as broader user characteristics or situational context—which is crucial for interactions requiring a comprehensive understanding of the user. As illustrated on the upper side of \autoref{idea}, an intelligent agent should consider the user’s health condition and recent schedule when planning weekend activities, rather than relying solely on activity-related memories. ii) The message grouping-based retrieval architecture can introduce additional retrieval noise. As shown on the lower side of \autoref{idea}, sub-optimal memory groups can compel the agent to retrieve information from all three groups to gather sufficient context for appropriate responses. These redundant retrievals diminish the effectiveness of the model’s responses, while also increasing latency and token consumption during LLM inference.

In this report, we propose the initial design of \method{}, a human-centric memory framework based on active user profiling. Unlike conventional approaches that merely store and group past interactions between users and agents for retrieval, \method{} actively extracts and updates user persona characteristics and event records from ongoing dialogs. This enables the agent to progressively refine its understanding of user attributes and historical experiences. We redefine personalized memory systems by treating each proactive user interaction as an opportunity for iterative user modeling. This approach effectively leverages both persona profiles and topic-related event records as contextual cues to support personalized responses. The main contributions of this work are as follows:

\begin{itemize}[leftmargin=*]


     \item We identify limitations in existing grouping-then-retrieval based semantic retrieval memory frameworks, notably their inadequate user understanding and restricted personalization abilities as they primarily depend on static historical interaction embeddings rather than constructing dynamic, multi-dimensional user contexts.

    \item We propose \method{}, a novel persona memory framework that utilizes dynamic user profiling and a hierarchical, user-centric retrieval strategy. Unlike approaches that rely solely on semantic retrieval of past messages, \method{} actively constructs and updates user profiles by accumulating knowledge from interaction histories.
    
    \item Extensive experiments on three persona-oriented tasks—i) persona-based open question answering, ii) persona-guided response selection and iii) persona-centric in-depth report generation—show that \method{} consistently improves performance in a variety of memory applications. Specifically, \method{} sets a new state-of-the-art in memory performance, achieving 51.76    \% on the public LoCoMo benchmark and 62.99\% on PERSONAMEM. By enabling dynamic user profiling and \textbf{interaction-time scaling}, \method{} allows LLM agents to continuously adapt to users' evolving needs, demonstrating strong potential for enhancing long-term human-AI interactions through more personalized responses.  
    

\end{itemize}

\section{Related Work}

\textbf{Agent Memory System} powered by large language models (LLMs) has gained significant popularity in recent years owing to their remarkable capabilities in task comprehension and execution \citep{shen2025mind,firat2023if,wu2024autogen,zhou2023agents,zhou2024agents2,zhu2025oagentsempiricalstudybuilding,li2025chainofagentsendtoendagentfoundation}. Nevertheless, these systems continue to grapple with the challenge of sustaining high-quality performance when incorporating historical experience across complex, long-duration scenarios \citep{kang2025memory,zhang2024survey}. To address this limitation, numerous agent memory enhancement frameworks have been proposed, which can be broadly classified into two categories: (i) approaches that fine-tune LLM parameters to enhance information memorization and utilization \citep{wei2025ai,zhou2025mem1,modarressi2024memllm}, and (ii) methods that employ sophisticated information organization and retrieval techniques within external memory systems to preserve LLMs' long-term capabilities~\citep{zhou2023recurrentgpt}. The latter approach has attracted considerable attention due to its plug-and-play nature, which eliminates the need for additional training costs. Furthermore, these methods significantly reduce the dependency of memory capacity on the LLMs' input window length. For example, Think-in-Memory (TiM) \citep{liu2023think} preserves the reasoning traces of LLMs across multiple dialogue rounds to alleviate response inconsistencies. A-Mem \citep{xu2025mem} organizes memory fragments into linked lists to improve retrieval performance. Grounded Memory \citep{ocker2025grounded} introduces vision language models (VLMs) to interpret consecutive audio frames, organizing these interpretations in a graph structure for subsequent retrieval. MemoryBank \citep{zhong2024memorybank} incorporates the Ebbinghaus Forgetting Curve theory to enable agents to forget and reinforce memories based on time elapsed and the relative significance of memory segments. MemGPT \citep{packer2023memgpt} and Memory OS \citep{kang2025memory} adopt an operating system-like architecture for memory organization and retrieval, employing mechanisms such as a first-in-first-out queue for working memory management. However, most existing systems overlook a critical aspect: how to dynamically and hierarchically establish connections between memory fragments to continuously update the agent's overall understanding of its environment. For instance, while A-Mem \citep{xu2025mem} and Memory OS \citep{kang2025memory} store semantically similar information in linked segments and retrieve the grouped data during response generation, their simple chunk-retrieval mechanisms, as illustrated in \autoref{idea}, often fail to equip agents with a comprehensive and in-depth understanding of users prior to interaction. Therefore, to bridge this gap, we propose a novel memory system, \method{}, based on active user profiling. The key difference between \method{} and previous memory systems is that its core task is to answer the questions: "What kind of person is this user? What has he or she experienced?" rather than merely grouping received user information for later retrieval.
We draw inspiration from the human brain's memory architecture and consequently redefine the three core components of an agent's memory: i). \textbf{Episodic Memory}, which is responsible for mapping a user's historical interaction cues to their corresponding situational contexts (e.g., mapping the cue "project deadline" to the specific episode where the user expressed stress and requested help with scheduling); ii). \textbf{Persona Memory}, which constructs and maintains a holistic profile of the user; iii). \textbf{Working Memory}, which is responsible for providing relevant contextual information to the current interaction. Together, these components work in concert to enable \method{} to build a deep, dynamic understanding of the user, powering more personalized and context-aware interactions. \autoref{idea} summarizes the key design features 
coherence of our approach in comparison with representative existing memory systems. We refer the conflict management as the process of memory systems to maintain the coherence of its stored information and agile user-centric user modeling as the process of timely iterating on the understanding of users.

\textbf{Personalized Agent.} While large language models (LLMs) serve as powerful assistants for a multitude of tasks, their effectiveness remains constrained without the ability to learn from and adapt to human preferences through personalization. A promising direction involves the development of persona agents—LLM-based systems deeply integrated with personal data to deliver responses aligned with user-specific needs \citep{li2024personal}. Meeting the growing demand for such personalized interactions requires methodologies that can continuously and accurately infer user characteristics from their interactions \citep{sun2024persona, magister2024way, eapen2023personalization,wang2024aipersonalifelongpersonalization}. Most prior work on persona-enhanced LLMs has focused on injecting user information through fine-tuning \citep{salemi2024optimization, eapen2023personalization,tan2024democratizing,zhu2025faithfulcontrollablepersonalizationcritiquepostedit} or direct retrieval from user traces of static user profiles that rely on a limited set of predefined attributes\citep{richardson2023integrating,sun2024persona,qiu2025measuring,wang2024aipersonalifelongpersonalization,tao2025personafeedbacklargescalehumanannotatedbenchmark}. However, these approaches face significant limitations in handling long-term, dynamic and evolving user preferences: fine-tuning requires computationally expensive retraining for each update, while direct retrieval lacks the capacity to synthesize longitudinal interaction patterns into coherent and evolving user profiles. In this work, we propose a persona memory system that dynamically organizes a user's interaction history into structured persona characteristics and experiential data, enabling more precise, adaptive, and personalized responses over time.

\begin{table}[t]
    \centering
    \caption{Comparison of design features across different systems}
    \vspace{-9pt}
    \resizebox{\linewidth}{!}{
    \begin{tabular}{l|ccccc}
    \toprule
    \textbf{System} & \textbf{\makecell{Conflict \\Management}} & \textbf{\makecell{Multi-Memory\\Components Design}} & \textbf{\makecell{Multi-Channel\\ Retrieval}} & \textbf{\makecell{Independent of \\Pre-chunking}} & \textbf{\makecell{Agile User-Centric\\ Modeling}} \\
    \midrule
        Mem0\cite{chhikara2025mem0}     & \textcolor[RGB]{0,120,0}{\ding{51}} & \textcolor{red}{\ding{55}} & \textcolor{red}{\ding{55}} & \textcolor[RGB]{0,120,0}{\ding{51}} & \textcolor{red}{\ding{55}} \\
        MemoryOS\cite{kang2025memory} & \textcolor[RGB]{0,120,0}{\ding{51}} & \textcolor[RGB]{0,120,0}{\ding{51}} & \textcolor[RGB]{0,120,0}{\ding{51}} & \textcolor{red}{\ding{55}} & \textcolor{red}{\ding{55}} \\
        A-Mem\cite{xu2025mem}    & \textcolor[RGB]{0,120,0}{\ding{51}} & \textcolor{red}{\ding{55}} & \textcolor{red}{\ding{55}} & \textcolor{red}{\ding{55}} & \textcolor{red}{\ding{55}} \\
        O-Mem    & \textcolor[RGB]{0,120,0}{\ding{51}} & \textcolor[RGB]{0,120,0}{\ding{51}} & \textcolor[RGB]{0,120,0}{\ding{51}} & \textcolor[RGB]{0,120,0}{\ding{51}} & \textcolor[RGB]{0,120,0}{\ding{51}} \\
    \bottomrule
    \end{tabular}}
    \label{comparison2}
    \end{table}

\section{Method}

When building the memory for O-Mem, we architect it to emulate a human-like memory model \cite{lai2025brain}. This is realized through three key properties: i).Long-Term Personality Modeling: It constructs a persistent and evolving user profile, mirroring the human ability to build a coherent understanding of others over time. ii).Dual-Context Awareness: It maintains both topical continuity (working memory) and associative, clue-triggered recall (episodic memory), enabling both coherent and precisely cued responses. iii).Structured, Multi-Stage Retrieval: It replaces a monolithic search with a structured process that orchestrates consultations with different memory types, resulting in more robust, transparent, and human-like reasoning.
As illustrated in \autoref{method}, \method{} continuously extracts and refines user profiles through ongoing interaction, building a semantic mapping between topics/clues and corresponding interaction scenarios. This enables dynamic, multi-faceted user understanding that supports powerful personalization. In this section, we first present the notation used in our work and the storage formats of different memory components (\autoref{notation}), followed by an explanation of how user interactions are encoded into the different memory components in \method{} (\autoref{memory indexing}), and finally describe the retrieval process across these memories (\autoref{memory construction}).

\subsection{Preliminary: Notation and Memory Components Storage Format}
\label{notation}

In this section, we define a user interaction, denoted as $U$, as a record of either explicit literal content (e.g., search queries) or implicit user behavior (e.g., taking a screenshot). Let $M_w$ be a dictionary that maps each clue word $w$ to the interactions in which it appears, and $M_t$ be a dictionary that maps each topic to its corresponding interactions. Additionally, $P_a$ denotes the list of persona attributes, and $P_f$ represents the list of persona fact events:
\[
\begin{aligned}
& \text{Let } \mathcal{U} = \{U_1, U_2, \dots, U_n\} \text{ be the set of user interactions.} \\
\end{aligned}
\]
\[
M_w[w] = \{ U \in \mathcal{U} \mid w \text{ appears in } U \}, \quad M_t[t] = \{ U \in \mathcal{U} \mid U \text{ is associated with topic } t \}.
\]
As illustrated in \autoref{method}, we model these components within a cognitive architecture: \( P_f \) and \( P_a \) constitute the \textbf{user persona memory}, which stores long-term, abstracted user knowledge; \textbf{the topic-interaction dictionary \( M_t \)} functions as \textbf{working memory}, capturing the topical context of the current interaction; and the \textbf{keyword-interaction dictionary \( M_w \)} serves as \textbf{episodic memory}, acting as an associative index that links salient clues to their originating interactions. Unlike the strict physiological definitions of working memory and episodic memory, \textbf{The definitions of agent working memory and episodic memory in \method{} are past interactions related to the current interaction topic and past interactions related to clues in the current interaction, respectively.} The semantic similarity function \( s(t_1, t_2) \) between two text segments \( t_1 \) and \( t_2 \), and the memory retrieval function \( F_{\text{Retrieval}} \) based on \( s \), are formally defined as follows:
\[
s(\mathbf{t_1}, \mathbf{t_2}) = \frac{\mathbf{f_e(t_1)} \cdot \mathbf{f_e(t_2)}}{\|\mathbf{f_e(t_1)}\| \|\mathbf{f_e(t_2)}\|}, \quad F_{\text{Retrieval}}(M \mid q) = \text{top-}k\big\{s(m, q) \mid m \in M\big\}
\]
where \( f_e(\cdot) \) denotes a text embedding function, \(\text{top-}k\) returns the \(k\) most similar items, and \( M \) refers to the memory component from which retrieval is performed.  
\definecolor{atomictangerine}{rgb}{1.0, 0.6, 0.4}
\definecolor{babyblue}{rgb}{0.54, 0.81, 0.94}
\definecolor{asparagus}{rgb}{0.53, 0.66, 0.42}

\begin{figure}[!t]
    \centering
\includegraphics[width=1\textwidth]{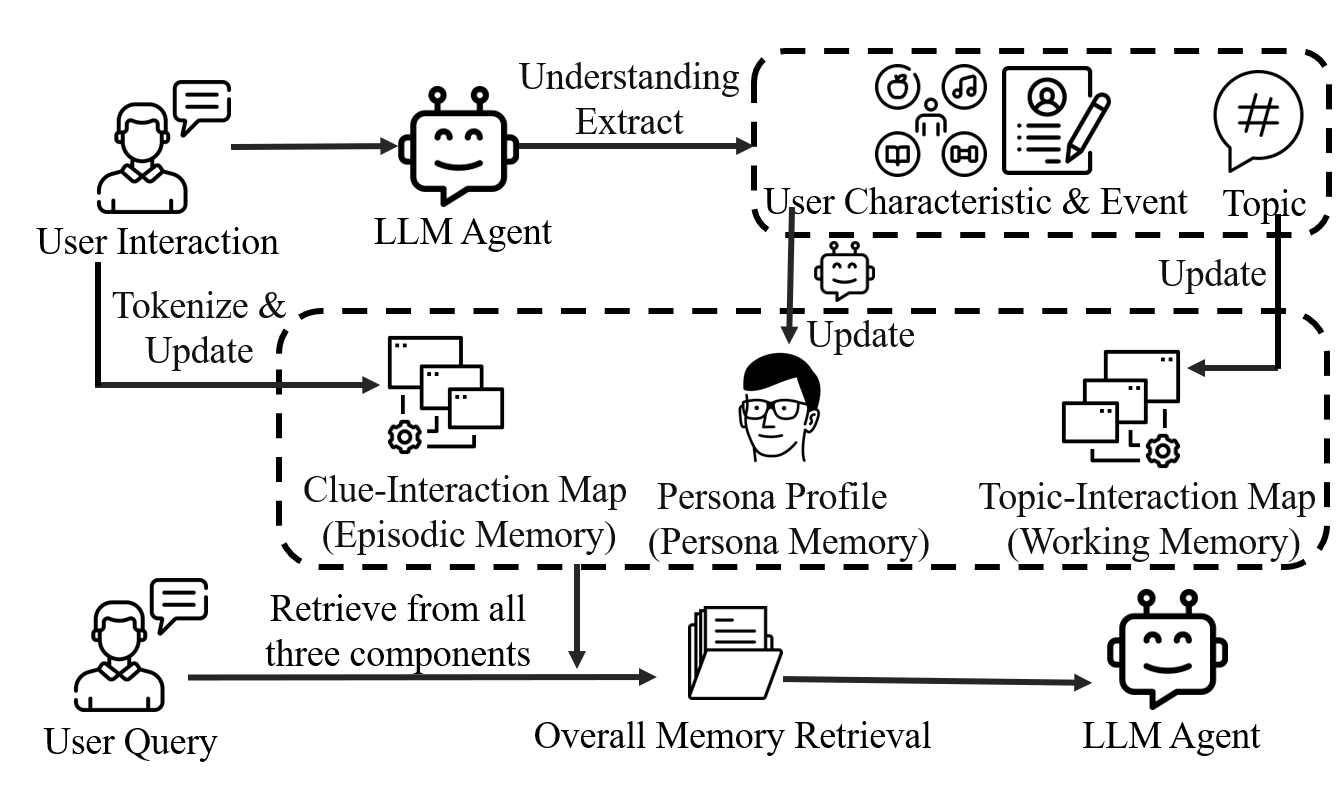}
        \caption{Top: The process of encoding user interactions into memory in \method{}. Different colors refers to different memory components. \method{} encodes a user interaction into memory by extracting and recording relevant user attributes and event data into persona memory,episodic memory, and working memory. Bottom: The memory retrieval process concerning one user interaction in \method{}. \method{} retrieves from all its three memory components concerning one new user query.}
    \label{method}
\end{figure}


\subsection{Memory Construction Process}\label{memory indexing}

There are three components of \method{}: i).Persona Memory: This component stores the user's long-term attributes and significant factual events in a structured profile. Its function is to enable Long-Term Personality Modeling, ensuring responses are consistently personalized by maintaining a persistent understanding of the user's identity.
ii).Working Memory: This component maintains a dynamic index mapping conversation topics to related interactions. Its function is to support dual-context awareness by preserving topical continuity, thereby ensuring dialogue coherence through relevant contextual grounding. iii).Episodic Memory: This component serves as an associative index linking specific clues or keywords to their original interactions. Its function is to enable associative, clue-triggered recall as part of dual-context awareness, allowing precise retrieval of specific events beyond mere semantic similarity. The three memory components employ distinct technical approaches tailored to their specific functions: persona memory uses LLM-based extraction to detect user attributes and events, with regulation through a decision process (Add/Ignore/Update) to maintain profile coherence. Working memory automatically indexes interactions under LLM-identified topics, relying primarily on accurate topic detection for regulation. Episodic memory tokenizes interactions to detect potential clues, regulated by a distinctiveness filter that prioritizes rare keywords. This statistics-driven approach enables precise associative recall. The differentiation in update mechanisms stems from their distinct purposes: persona memory requires careful curation, while working and episodic memories benefit from automated indexing for efficiency.

Given the $i$-th user interaction $u_i$, \method{} first extracts its topic $t_i$, revealed user attribute $a_i$ and past event $e_i$ with language model $\mathcal{L}$:
{\small
\begin{equation}
(t_i, a_i, e_i) = \mathcal{L}(u_i)
\end{equation}
}The clue interaction map $M_w$ and topic interaction map $M_t$ are updated by increasing the count for each word in $u_i$ and $t_i$ :
{\small
\begin{equation}
M_t^{(i+1)}[t_{i}] \gets M_t^{(i)}[t_{i}] \cup \{i\}, \quad
M_w^{(i+1)}[w_j] \gets M_w^{(i)}[w_j] \cup \{i\}, \quad \forall\, w_j \in \mathcal{T}(u_{(i)})
\end{equation}
}where $\mathcal{T}(u_{(i)}) = \{w_1, w_2, \dots, w_n\}$ represents the tokenized words from the $i$-th user interaction. For $e_i$, $\mathcal{L}$ generates an memory management operation regarding its integration with the existing persona fact event list $P_f$:
{\small
\begin{equation}
\text{Op}(e_i) \gets \mathcal{L}\big(e_i, P_f\big) \in \{\text{Add,Ignore,Update}\}, \quad P_f \gets \text{ApplyOp}(P_f,e_i,\text{Op}(e_i))
\end{equation}
}

where $\text{Op}$ refers to the operation decision from $\mathcal{L}$ and $\text{ApplyOp}$ refers to the function that executes this operation. During our observation, we identified that similar attributes from the same user frequently recur across different interactions (e.g., users always mention their hobbies repeatedly). To better organize these extracted attributes, we propose an LLM-augmented nearest neighbor clustering method: 
{\small
\begin{gather}
\text{Op}(a_i) \gets \mathcal{L}(a_i, P_a^{t}) \in \{\text{Add, Ignore, Update}\},\quad P_a^{t} \gets \text{ApplyOp}(P_a^{t}, a_i, \text{Op}(a_i)) \\
\text{NN}(a_i) = \argmin_{{a}_l \in P_a^{t}, l \neq i}\left(1 - s(a_i, a_l)\right) \\
G = (V, E),\ V = \{a_1, \dots, a_K\},\ E = \{(a_l, \text{NN}(a_l)) \mid a_l\in P_a^{t}\} \\
\mathcal{B} = \{B_1, \dots, B_M\} = \text{ConnectedComponents}(G),\quad P_a = \bigcup_{m=1}^M \mathcal{L}(B_m)
\end{gather}}
where $\mathrm{NN}(a_i)$ denotes the nearest neighbor of attribute $\mathbf{a}_i$ based on the similarity function $s(\cdot,\cdot)$; $G = (V, E)$ represents the nearest-neighbor graph constructed from the temporary attribute list $P_a^{t}$, with vertices corresponding to attributes and edges connecting each attribute to its nearest neighbor; $K$ refers to the total number of attributes in $P_a^{t}$, $\mathcal{B} = {B_1, \dots, B_M}$ is the set of connected components obtained from $G$ via connected component analysis; and the final attribute set $P_a$ is obtained by applying the large language model $\mathcal{L}$ to analyze the aggregated attributes within each connected component.

\subsection{Memory Retrieval Process}
\label{memory construction}

\method{} employs a parallel retrieval strategy across all three memory components. The system simultaneously queries the Persona Memory for user profile context, Working Memory for topical context, and Episodic Memory for clue-triggered events. The retrieved information from all memory components is then aggregated and processed by the language model to generate the final response. This parallel approach ensures comprehensive context integration while maintaining the distinct advantages of each component. For each user interaction $u_{i}$, \method{} conducts retrieval from the user's persona memory, episodic memory, and working memory. We introduce their retrieval process separately. 

\noindent \textbf{Working Memory Retrieval.} we define the retrieval process of working memory as:
{\small
\begin{equation}
R_{working} = \bigcup_{t \in \hat{T}} M_t[t], \quad \text{where} \quad \hat{T} = F_{\text{Retrieve}}(\mathcal{K}(M_t), u_{i})
\end{equation}}where $M_t$ is the mapping from topics to their corresponding interactions, $\mathcal{K}(M_t)$ denotes the set of topics in $M_t$, $u_i$ is the current interaction, $F_{Retrieve}$ retrieves the most relevant topics $\hat{T}$ for $u_i$ from $\mathcal{K}(M_t)$, and $R_{working}$ is the set of interactions in $M_t$ corresponding to these relevant topics.

\noindent \textbf{Episodic Memory Retrieval.} We define the retrieval process of episodic memory as follows. The episodic memory is structured as a word-to-interactions mapping dictionary $M_w$ ($M_w: w \rightarrow \{i\}$), which maps words to the sets of past interactions (memory entries) in which they appear. That is, for a word $w$, $M_w[w]$ yields all past interactions containing $w$. Given the current user interaction $u_i$, the retrieval process is: (1) \textbf{Tokenization:} Tokenize the utterance into a sequence of words: $W = \text{Tokenize}(u_i)$. (2) \textbf{Clue Selection:} Calculate the clue selection score for each word $w \in W$ with respect to the clue-interaction map $M_w$. The word with the highest score is selected as the target clue $\hat{w}$:
{\small
\begin{equation}
    \hat{w} = \underset{w \in W}{\arg\max} \, \text{Score}(w, M_w)
\end{equation}
\begin{equation}
    \text{Score}(w, M_w) = \frac{1}{df_w}
\end{equation}}where $df_w$ is the number of past interactions in $M_w$ that contain the word $w$ (i.e., $df_w = |M_w[w]|$). The set of episodic memory interactions associated with the clue $\hat{w}$ is then retrieved as: $R_{episodic} = M_w[\hat{w}]$.

\noindent \textbf{Persona Memory Retrieval.} We define the persona retrieval process as:
{\small
\begin{equation}
R_{persona}=F_{\text{Retrieval}}(P_{f},u_{i})\oplus F_{\text{Retrieval}}(P_{a},u_{i})
\end{equation}}where $P_{f}$ refers to the persona facts, $P_{a}$ refers to the persona attributes, $u_i$ is the current user interaction, $\oplus$ denotes the concatenation operation, and $R_{persona}$ refers to the retrieved persona information.

\noindent \textbf{Overall Memory Retrieval.}
We define the overall retrieval and final response as:
\begin{align}
R = R_{\text{working}} \oplus  R_{\text{episodic}} \oplus   R_{\text{persona}},
O = \mathcal{L}\big(R,u_{i}\big)
\end{align}
where $R$ represents the overall retrieved memory content, $O$ represents the final response generated by the language model $\mathcal{L}$ based on the current user interaction $u_i$ and the retrieved memories.


\section{Experiment}

\textbf{Datasets and Evaluation Metrics.} 
We evaluate our method on three benchmarks: 
\textbf{LoCoMo}~\citep{maharana2024evaluating}, \textbf{PERSONAMEM}~\citep{jiang2025know}, and \textbf{Personalized Deep Research Bench~\citep{liang2025towards}}.

The \textbf{LoCoMo} benchmark features extended dialogues averaging 300 turns across four memory challenge types: Single-hop, Multi-hop, Temporal, and Open-domain. 
The \textbf{PERSONAMEM} dataset contains user-LLM conversations spanning 15 diverse topics. 
To address the need for evaluating personalized long-text generation, we introduce \textbf{Personalized Deep Research Bench}, a benchmark simulating real-world deep research scenarios \cite{liang2025towards}. Unlike existing datasets, Personalized Deep Research Bench comprises 50 deep research queries derived from multi-round conversations between 25 real users and LLMs, requiring nuanced understanding of individual user characteristics. It is built upon a subset of a persona deep research dataset collected from real users through commercial applications but is specifically repurposed and curated by the dataset construction committee for assessing memory system\footnote{The full original Personalized Deep Research Bench benchmark has been released to the community comprises the entire initial query set, including the 50 highly personalized memory-related queries selected for this study as well as the broader collection, which, despite being less discriminative for memory personalization, remains a valuable asset for future research. Specifically, the data construction committee removed some queries that were too difficult or too easy for the memory system to answer from the original deep research dataset, making it more cost-effective.} For evaluation, we employ: 

\begin{itemize}
    \item \textbf{LoCoMo}: F1 and BLEU-1 scores following the standard protocol;
    \item \textbf{PERSONAMEM}: Accuracy for multiple-choice questions;
    \item \textbf{Personalized Deep Research Bench}: Goal Alignment and Content Alignment scores, measuring adherence to user characteristics and expectations via LLM-as-a-judge. 
\end{itemize}

\textbf{Compared Baseslines.}
Our method is compared with: 
(i) open-source memory frameworks: A-Mem~\citep{xu2025mem}, MemoryOS~\citep{kang2025memory}, Mem0~\citep{chhikara2025mem0}, and LangMem~\citep{Langchain-Ai}; and 
(ii) commercial/proprietary frameworks: ZEP~\citep{rasmussen2025zep}, Memos~\citep{li2025memos}, and OpenAI~\citep{Memory-OpenAI}. 
\textbf{Due to budget constraints and licensing costs}, we report results from original publications for commercial frameworks. Mem0 was evaluated using its open-source version due to cost and accessibility.  

\textbf{Implementation Details.}
We use all-MiniLM-L6-v2 \citep{all_MiniLM_L6_v2} as embedding model in \method{} to calculate similarities. All of our experiments are conducted on two A800 GPUs. The choice of language models across datasets was informed by computational budget. A comparative analysis using both GPT-4.1 and GPT-4o-mini was performed on the LoCoMo benchmark. For the remaining datasets (PERSONAMEM and Personalized Deep Research Bench), only the larger GPT-4.1 model was used. Due to resource constraints, the experiments were conducted on public shared servers. The hardware resources in cloud environments (e.g., GPU computational capacity, network latency) may exhibit inherent volatility, meaning we cannot guarantee identical latency and computational states across experimental runs. The generative nature of Large Language Models (LLMs) is inherently stochastic. Given the significant economic cost associated with API calls, it was not feasible to perform multiple repeated experiments to obtain statistical summaries (e.g., mean and standard deviation). Furthermore, in pursuit of evaluation fairness, we deliberately avoided fixing the random seed to prevent any potential—even unintentional—"seed cherry-picking," thereby ensuring the reported result represents a single sample from the distribution of possible outcomes. Therefore, we emphasize that the primary contribution of this work is to demonstrate the feasibility and fundamental trends of the proposed method, rather than to provide a highly precise performance benchmark under controlled conditions. We encourage readers to focus their evaluation on the relative trends within the same unstable environment and to interpret the absolute performance values with caution. Owing to regional constraints, access to the GPT model was facilitated through an intermediary API provider. However, the reliability of this service proved erratic, manifesting as frequent request failures and malformed responses. These issues, which were exacerbated during large-scale data construction, introduced significant uncertainty into our estimates of computational cost and time expenditure. Consequently, this variability precluded us from reporting precise values for these metrics.
Mem0 was evaluated using its open-source version due to cost and accessibility. For efficiency, the GPT-4o results for baselines (excluding Mem0) are adopted from existing literature.

\begin{table}[!t]
\small
\centering
\caption{Performance comparison using different LLMs on LoCoMo with best scores highlighted.}
\begin{tabular}{@{}cc*{10}{c}@{}}
\toprule
\multicolumn{1}{c}{LLM} & \multicolumn{1}{c}{Method} & \multicolumn{2}{c}{Cat1: Multi-hop} & \multicolumn{2}{c}{Cat2: Temporal} & \multicolumn{2}{c}{Cat3: Open} & \multicolumn{2}{c}{Cat4: Single-hop} & \multicolumn{2}{c}{Average} \\
    &        & F1 & B1 & F1 & B1 & F1 & B1 & F1 & B1 & F1 & B1 \\ 
\midrule
\multirow{5}{*}{GPT-4.1} 
& LangMem & 41.11 & 32.09 & 53.67 & 46.22 & \textbf{33.38} & \textbf{27.26} & 51.13 & 44.22 & 48.72 & 41.36 \\
& Mem0\tnote{*} & 30.45 & 22.15 & 10.69 & 9.21 & 16.75 & 11.34 & 30.32 & 25.82 & 25.40 & 20.78 \\
& MemoryOS & 29.25 & 20.79 & 37.73 & 33.17 & 22.70 & 18.65 & 43.85 & 38.72 & 38.58 & 33.03 \\
& A-Mem & 29.29 & 21.47 & 33.12 & 28.50 & 15.41 & 12.34 & 37.64 & 32.88 & 33.78 & 28.60 \\  

& \cellcolor{DarkSeaGreen3!40}Ours & \cellcolor{DarkSeaGreen3!40}\textbf{42.64} & \cellcolor{DarkSeaGreen3!40}\textbf{34.08} & \cellcolor{DarkSeaGreen3!40}\textbf{57.48} & \cellcolor{DarkSeaGreen3!40}\textbf{49.76} & \cellcolor{DarkSeaGreen3!40}30.58 & \cellcolor{DarkSeaGreen3!40}25.69 & \cellcolor{DarkSeaGreen3!40}\textbf{54.89} & \cellcolor{DarkSeaGreen3!40}\textbf{48.98} & \cellcolor{DarkSeaGreen3!40}\textbf{51.67} & \cellcolor{DarkSeaGreen3!40}\textbf{44.96} \\ 

\midrule

\multirow{8}{*}{GPT-4o-mini}
& LangMem & 36.03 & 27.22 & 38.10 & 32.23 & 29.79 & 23.17 & 41.72 & 35.61 & 39.18 & 32.59 \\
& Mem0 & 17.19 & 12.06 & 3.59 & 3.37 & 12.24 & 8.57 & 12.74 & 10.62 & 11.62 & 9.24 \\
& ZEP & 23.14 & 14.96 & 17.59 & 14.57 & 19.76 & 13.17 & 32.49 & 27.38 & 26.88 & 21.55 \\
& MemoryOS & 41.15 & 30.76 & 20.02 & 16.52 & \textbf{48.62} & \textbf{42.99} & 35.27 & 25.22 & 34.00 & 25.53 \\
& OpenAI & 33.10 & 23.84 & 23.90 & 18.25 & 17.19 & 11.04 & 36.96 & 30.72 & 32.30 & 25.63 \\
& A-Mem & 33.23 & 29.11 & 8.04 & 7.81 & 34.13 & 27.73 & 22.61 & 15.25 & 22.24 & 17.02 \\
& MEMOS & 35.57 & 26.71 & \textbf{53.67} & \textbf{46.37} & 29.64 & 22.40 & 45.55 & 38.32 & 44.42 & 36.88 \\

& \cellcolor{DarkSeaGreen3!40}Ours & \cellcolor{DarkSeaGreen3!40}\textbf{44.17} & \cellcolor{DarkSeaGreen3!40}\textbf{34.78} & \cellcolor{DarkSeaGreen3!40}53.54 & \cellcolor{DarkSeaGreen3!40}45.65 & \cellcolor{DarkSeaGreen3!40}25.24 & \cellcolor{DarkSeaGreen3!40}19.22 & \cellcolor{DarkSeaGreen3!40}\textbf{54.53} & \cellcolor{DarkSeaGreen3!40}\textbf{48.33} & \cellcolor{DarkSeaGreen3!40}\textbf{50.60} & \cellcolor{DarkSeaGreen3!40}\textbf{43.48} \\
\bottomrule
\end{tabular}
\label{main_result_1}
\end{table}

\begin{table}[!t]
\small
\centering
\caption{Performance comparison on PERSONAMEM with GPT-4.1.}
\scalebox{0.95}{
\begin{tabular}{cccccccc}
\hline 
Method         & \begin{tabular}[c]{@{}c@{}}Recall user \\shared facts\end{tabular} & \begin{tabular}[c]{@{}c@{}}Suggest new\\ ideas\end{tabular} & \begin{tabular}[c]{@{}c@{}}Track full \\ preference \\ evolution\end{tabular} & \begin{tabular}[c]{@{}c@{}}Revisit reasons \\ behind\\ preference updates\end{tabular} & \begin{tabular}[c]{@{}c@{}}Provide preference-\\ aligned \\ recommendations\end{tabular} & \multicolumn{1}{l}{\begin{tabular}[c]{@{}l@{}}Generalize to \\ new scenarios\end{tabular}} & Average        \\
\hline 
LangMem     & 31.29                                                     & 24.73                                            & 53.24                                                                & 81.82                                                                         & 40.00                                                                           & 8.77                                                                              & 42.61 \\
Mem0           & 32.13                                                     & 15.05                                              & 54.68                                                                & 80.81                                                                         & 52.73                                                                           & 57.89                                                                             & 46.86 \\
A-Mem & 63.01                                                     & \textbf{27.96}                                             & 54.68                                                                & 85.86                                                                         & 69.09                                                                           & 57.89                                                                             & 59.42 \\
Memory OS &      \textbf{72.72}                                                &   17.20                                            &   58.27                                                            & 78.79                                                                        &   \textbf{72.72}                                                                        & 56.14                                                                              &  58.74\\

\rowcolor{DarkSeaGreen3!40}\method{}           & 67.81                                                     & 21.51                                              & \textbf{61.15}                                                                & \textbf{89.90}                                                                         & 65.45                                                                           & \textbf{73.68}                                                                             & \textbf{62.99}  \\
\hline 
\end{tabular}
}
\label{main_result_2}
\end{table}

\begin{table}[!htbp]
\small
\centering
\caption{Performance comparison on Personalized Deep Research Bench with GPT-4.1.}
\scalebox{1.00}{
\begin{tabular}{ccccc}
\hline 
Method & 
\multicolumn{1}{p{2cm}}{\centering Goal Alignment} & 
\multicolumn{1}{p{1.5cm}}{\centering Content Alignment} & 
Average \\
\hline 
Mem0 &  37.32 & 35.54 & 36.43\\
Memory OS & 40.60 & 39.67 & 40.14 \\
\rowcolor{DarkSeaGreen3!40}\method{} & \textbf{44.69} &  \textbf{44.29} & \textbf{44.49} \\
\hline 
\end{tabular}}
\label{main_result_3}
\begin{flushleft}  
\small
\justifying 
\end{flushleft}
\end{table}



\textbf{Performance Comparison.} The experimental results in three benchmark datasets are separately shown in \autoref{main_result_1},\autoref{main_result_2},and \autoref{main_result_3}. Due to limited access to ZEP\citep{rasmussen2025zep}, Memos\citep{li2025memos}, and OpenAI memory\citep{Memory-OpenAI}, we only report their performance reported in their work using GPT-4o-mini. For Personalized Deep Research Bench benchmark dataset, we only compare our method with mem0\citep{chhikara2025mem0} and MemoryOS\citep{kang2025memory} for cost efficiency. \method{} demonstrates superior performance compared to all baselines across three benchmark datasets. The performance advantage is more pronounced in complex reasoning tasks. As shown in \autoref{main_result_1}, on the comprehensive LoCoMo benchmark, \method{} achieves the highest average F1 scores of 51.67\% with GPT-4.1 and 50.60\% with GPT-4o-mini, outperforming the strongest baselines by significant margins (2.95\% and 6.18\% absolute improvements, respectively). The performance advantage is particularly pronounced in complex reasoning tasks. For Temporal reasoning, \method{} achieves F1 scores of 57.48\% (GPT-4.1) and 53.54\% (GPT-4o-mini), substantially outperforming all baselines. This indicates that our memory management mechanism effectively handles temporal dependencies and sequential information, which is crucial for maintaining coherent long-term conversations. \autoref{main_result_2} further demonstrates \method{}'s effectiveness in personalized interaction scenarios on the PERSONAMEM dataset. \method{} achieves an average accuracy of 62.99\%, exceeding the closest competitor (A-Mem at 59.42\%) by 3.57\%. Notably, \method{} excels in challenging tasks such as "Generalize to new scenarios" (73.68\%) and "Revisit reasons behind preference updates" (89.90\%), highlighting its robust capability in understanding and adapting to evolving user preferences. The superiority of \method{} is consistently validated on our newly introduced Personalized Deep Research Bench dataset (\autoref{main_result_3}), where it achieves an average alignment score of 44.49\%, significantly higher than Mem0 (36.43\%). This 8.06\% improvement demonstrates our method's practical utility in real-world personalized deep research scenarios that require nuanced understanding of user characteristics. A fair comparison was conducted by generating all deep research reports through the centralized sonar-deep-research service \citep{Sonar_Deep_Research}, leveraging retrievals from each method's individual memory system.

\section{Discussion}

\textbf{Rethinking the Value of Memory Systems.} Do we truly need meticulously designed, complex memory systems? Most existing approaches adhere to a common paradigm: during retrieval, systems access processed user interactions rather than raw historical data. This design is largely driven by increasingly stringent privacy regulations worldwide \citep{voigt2017eu, calzada2022citizens, hosseini2024bilingual}. By relying on abstracted user data, memory systems help AI companies \textbf{mitigate legal risks} while maintaining personalization capabilities. However, this abstraction comes at a significant cost: \textbf{the irreversible loss of information fidelity and contextual nuance.} 
For instance, a detailed user statement such as \textit{``admiring a specific lamp and rug in a downtown antique store last Saturday''} may be compressed into a structured preference like \textit{``[User] likes vintage home decor.''} While efficient, this compression sacrifices granular details—specific objects, locations, and temporal context—that are crucial for precise and contextually relevant interactions.

To quantify this trade-off, we compare the performance of direct retrieval-augmented generation (RAG) overall raw interactions\footnote{This includes all original interactions. To avoid missing contextual information in direct retrieval, the responses from the agent are also retrieved, which is different from our work that focuses on modeling user interactions.} with \method{}. As shown in \autoref{tab:rag}, direct RAG achieves competitive performance (50.25 vs. 51.67 F1 score) despite its conceptual simplicity, though at a substantially higher computational cost (2.6K vs. 1.5K tokens). Notably, when compared to the results in \autoref{main_result_1}, direct RAG with access to complete interaction history achieves competitive performance, highlighting the fundamental value of preserving raw interaction data. However, this comes at prohibitive computational costs that limit practical deployment. \method{} addresses this critical limitation by achieving comparable performance with significantly reduced overhead, positioning it as a computationally efficient alternative that balances performance with practicality.

As depicted in \autoref{tab:rag}, we also evaluate the practical deployment advantages of \method{} by measuring average peak retrieval memory overhead \citep{liu2024retrievalattention} per response. \method{} achieves a significant 30.6\% reduction in peak memory overhead (from 33.16 MB to 22.99 MB), substantially relaxing hardware constraints for large-scale personalized inference.  While direct RAG processes complete interaction histories verbatim, O-Mem maintains distilled representations that preserve semantic essence while drastically reducing sequence length. This design choice yields substantial computational memory benefits. For response latency, \method{} demonstrates a 41.1\% reduction in delay compared to direct RAG (from 4.01 seconds to 2.36 seconds), highlighting its efficiency for real-time applications.  

\begin{table}[t]
\small
\centering
\caption{Performance and efficiency comparison between direct retrieval from complete raw interaction history (Direct RAG) and \method{}.}
\scalebox{1.00}{
\begin{tabular}{ccccc}
\hline
Method & F1 (\%) & Avg. Token Cost & Peak Memory Overhead (MB)\tablefootnote{*} & Delay (s) \\
\hline
Direct RAG & 50.25 & 2.6K & 33.16 & 4.01 \\ 
\method{} & 51.67 & 1.5K & 22.99 & 2.36 \\ 
\hline 
\end{tabular}}
\label{tab:rag}
\begin{flushleft}  
\small
\justifying 
\textsuperscript{*}{For a fair comparison, the reported overhead for both RAG and \method{} is calculated as the peak GPU memory usage minus the fixed memory allocated by the same embedding model used in our main experiments.}  
\end{flushleft}
\end{table}



\begin{figure}
    \centering
\includegraphics[width=1\linewidth]{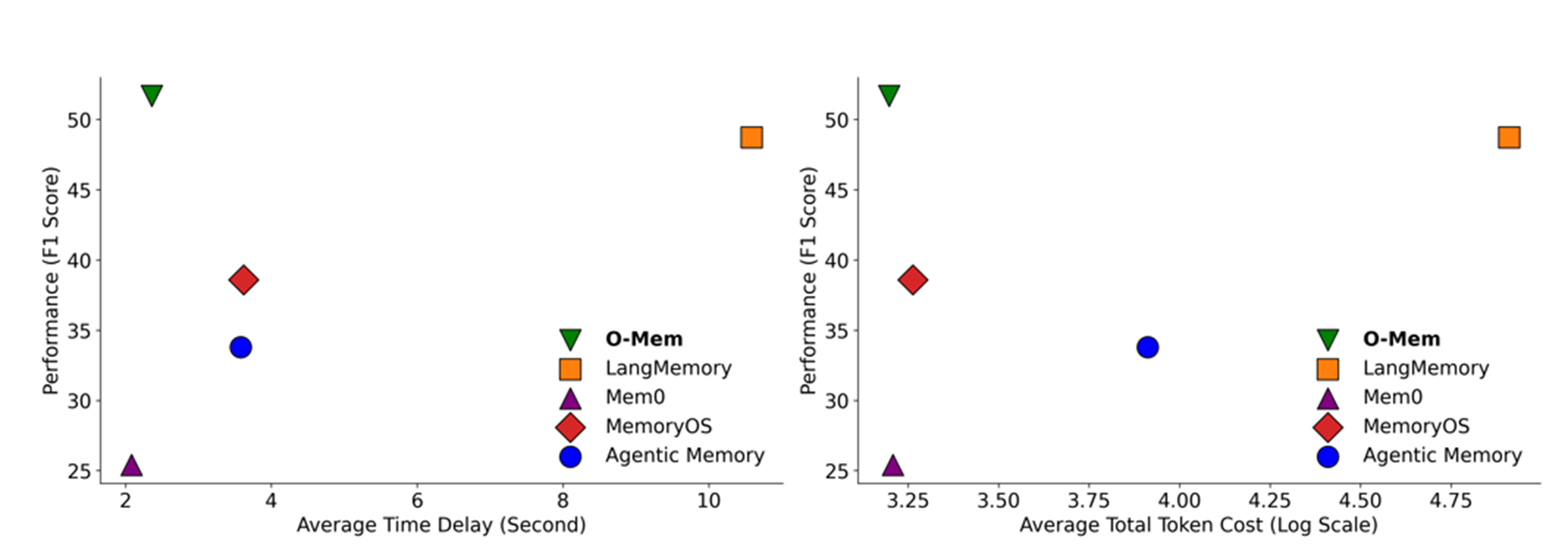}

\caption{Trade-off between performance and efficiency of different memory frameworks. The left panel (a) compares the average latency per interaction. The MemoryOS latency was evaluated using FAISS-CPU due to compatibility issues on our computing platforms, thus representing a conservative estimate of its latency. (b) The right panel compares the average computational cost (in tokens) per interaction. Results demonstrate that \method{} achieves a Pareto-optimal solution in both efficiency and overall performance. Note that we only performed token-control experiments on Locomo's GPT-4.1 experiments. We did not control tokens for the experiments on GPT-4o-mini and the other two datasets. }
    \label{time and cost efficiency}
\end{figure}

\textbf{Efficiency Analysis.} We evaluate the efficiency of \method{} by measuring the average token consumption and latency per response on the LoCoMo benchmark. The results, presented in \autoref{time and cost efficiency}, substantiate that \method{} achieves a superior balance between efficiency and effectiveness. Compared to the highest-performing baseline, LangMem (48.72 F1), \method{} (51.67 F1) reduces token consumption by \textbf{94\%} (from 80K to 1.5K tokens) and latency by \textbf{80\%} (from 10.8s to 2.4s), while delivering superior performance. Against the \textbf{second-best performing} baseline, MemoryOS (38.58 F1), \method{} not only secures a \textbf{34\%} higher F1 score but also reduces latency by \textbf{34\%} (2.4s vs. 3.6s). These results unequivocally demonstrate that \method{} sets a new Pareto frontier for efficient and effective memory systems.


The efficiency advantage of \method{} stems from two key design choices: The first is the independence of its retrieval operations across the three memory components. Unlike sequential architectures (e.g., A-men) that rely on a cascade of coarse-to-fine stages, \method{} performs a one-time, concurrent retrieval across all three memory paths. Secondly, the retrieval mechanism of \method{} utilizes user persona information, which, as opposed to raw user interaction records, typically contains less noise, thereby enhancing the cost-effectiveness of token usage. \method{} also achieves a substantial reduction in \textbf{storage footprint}, requiring only nearly \textbf{3 MB per user}--- much less than the nearly \textbf{30 MB per user} consumed by Memory OS. This storage efficiency stems from our topic/keyword-based mapping design, which utilizes a lightweight index, in contrast to Memory OS which must store dense vector mappings for each memory chunk. Furthermore, \method{} employs a radically simplified inference pipeline. Each response is generated through \textbf{only one} LLM invocation (three times for LangMemory). This streamlined workflow enables \method{} to achieve superior efficiency with minimal reponse latency and computational expense.

\textbf{Memory Component Analysis}
To quantify the contribution of each core module in our framework, we performed an ablation study on all three memory components—Persona Memory (PM), Episodic Memory (EM), and Working Memory (WM)—using the LoCoMo benchmark. The results are summarized in \autoref{tab:component analysis}. As indicated in the first three rows of the table, each module individually contributes to improved overall performance.
However, \textbf{such performance gains could be partially attributed to the increased volume of retrieved information}, which leads to longer retrieval sequences and higher token consumption during response generation. This trade-off between performance and efficiency has often been overlooked in prior ablation studies of memory-augmented systems. To definitively isolate this confound, we conducted a token-controlled ablation study (Rows~4--5 in \autoref{tab:component analysis}), wherein the total token budget for each ablated configuration was fixed at 1.5K tokens to match that of the full \method{} framework (WM+EM+PM).The clear performance gradient under a fixed token budget provides conclusive evidence that the performance gains are attributable to the \textit{quality and relevance} of the information retrieved by each module, not merely to an increase in context. This finding confirms that each independent memory module of \method{} effectively captures distinct and complementary aspects of the interactions.

\begin{table}[t]
\caption{Ablation study on different components of \method{} using the LoCoMo benchmark dataset.}
\small
\centering
\begin{tabular}{cccc}
\hline
Memory Configuration & F1 (\%) & Bleu-1 (\%) & Total Tokens \\ 
\hline
WM only & 44.03 & 38.05 & 1.3K \\ 
WM + EM & 49.62 & 43.18 & 1.4K \\ 
WM + EM + PM & 51.67 & 44.96 & 1.5K \\ 
WM + EM (token-controlled) & 50.10 & 43.27 & 1.5K \\ 
WM only (token-controlled) & 46.07 & 39.95 & 1.5K \\ 
\hline
\end{tabular}
\label{tab:component analysis}
\begin{flushleft}  
\small
\justifying 
\end{flushleft}
\end{table}

\begin{minipage}{\textwidth}
  \begin{minipage}[t]{0.40\textwidth}
    \begin{figure}[H]
      \centering
      \includegraphics[width=0.8\linewidth]
      {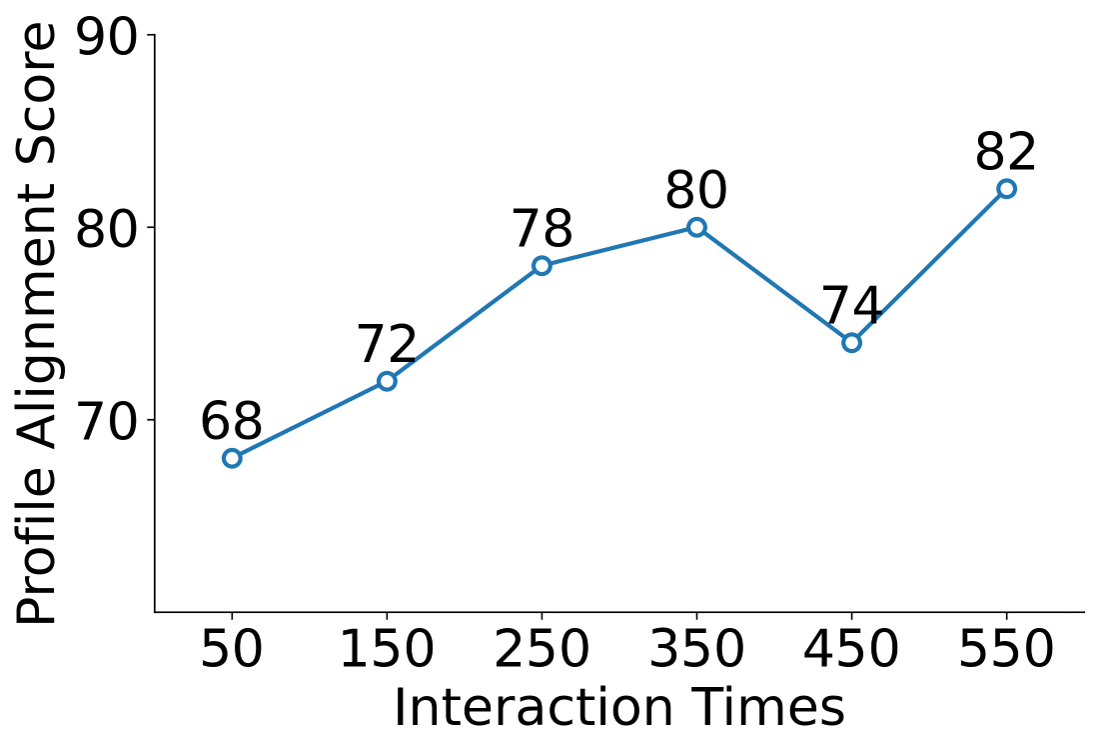}
      \caption{Memory profile alignment dynamics during memory-time Scaling. More interactions lead to more concise user understanding from \method{}.}
      \label{fig:profile-interaction}
    \end{figure}
  \end{minipage}%
  \hspace{0.2cm}
  \begin{minipage}[t]{0.47\textwidth}
    \begin{table}[H]
      \centering
      \small 
      \begin{tabular}{ccc}
        \hline
        \makecell{Memory\\Configuration} & \makecell{Average\\Performance} & \makecell{Average Retrieval \\Length (Chars)} \\
        \hline
        \method{}& 44.49 &6499  \\
        \makecell{\method{} w/o \\ Attributes} & 42.14 &28555  \\
        \hline
      \end{tabular}
      \caption{Ablation study on the impact of persona attributes, evaluated on the Personalized Deep Research Bench. Removing attributes from \method{} not only causes a performance drop but, more notably, leads to a substantial increase in retrieval length, indicating that attributes are crucial for precise memory filtering. This demonstrates the effectiveness and efficiency gains brought by user attribute extraction.}
      \label{tab:performance-comparison}
    \end{table}
  \end{minipage}
\end{minipage}

\textbf{Memory-Time Scaling for User Understanding.} We conduct a systematic evaluation of \method{}'s user understanding capability by examining how it \textbf{scales with the number of interactions} through two key analyses: (1) verifying the accuracy of persona attributes extracted from interaction data, and (2) assessing the practical utility of these attributes in personalizing the agent's responses. First, to evaluate the \textbf{scaling of extraction accuracy}, we collect persona attributes inferred by \method{} from a single user's dialogue history \textbf{across increasing interaction counts}. These extracted attributes are then compared against the user's ground-truth profile using an LLM-as-judge scoring mechanism \citep{li2024llms} to measure alignment. To evaluate the fidelity of the user profiles extracted by \method{}, we have designed structured prompts that direct a large language model to compare the extracted profiles against the ground-truth user profiles and assign a consistency score. We acknowledge the inherent limitations of this method in achieving perfect objectivity. Its primary objective, however, is not to provide a precise measurement but to fundamentally assess whether \method{} can construct a dynamic and increasingly accurate user portrait as the number of interactions scales. As shown in Figure~\ref{fig:profile-interaction}, the extracted persona attributes gradually converge toward the ground-truth profile \textbf{as interactions scale}, demonstrating that \method{} effectively refines its user understanding \textbf{through this scaling process}. Second, to measure the practical impact of persona attributes, we compare \method{} with and without access to persona attributes on the Personalized Deep Research Bench dataset. Results in Table~\ref{tab:performance-comparison} show that incorporating persona attributes yields a significant improvement in response personalization (average performance increases from 42.14 to 44.49) while substantially reducing the retrieval length (from 28,555 to 6,499 characters). These results demonstrate that \method{}’s ability to extract and leverage user attributes \textbf{through scaled interactions} substantially enhances its performance in complex personalized text generation tasks, achieving stronger personalization with improved efficiency.

\section{Conclusion}
In this paper, we propose \method{}, a novel memory framework that enhances long-term human-AI interaction through dynamic user profiling and hierarchical memory retrieval. Unlike conventional approaches that rely solely on semantic retrieval of past messages, \method{} actively constructs and refines user profiles from ongoing interactions. This approach effectively addresses the key limitations of conventional methods in maintaining long-term, consistent user context. Extensive experiments on three personalized benchmarks demonstrate that \method{} achieves state-of-the-art performance while reducing token consumption by 94\% and inference latency by 80\% compared to its closest competitor, highlighting its superior efficiency. The proposed framework provides an effective solution for complex personalized text generation tasks, enabling LLM agents to deliver more coherent and contextually appropriate responses. Our work opens up promising directions for developing more efficient and human-like personalized AI assistants in the future.

\newpage
\bibliographystyle{plainnat}
\bibliography{cite}
\newpage
\section{Contributions}

\textbf{Core Contributors}
\begin{tasks}[
            style=itemize,
            label-width=1em,
            column-sep=3.5em,
            before-skip=1.5ex,
            after-item-skip=1.5ex,
            ](2)
    \task Piaohong Wang\textsuperscript{*}
    \task Motong Tian\textsuperscript{*}
\end{tasks}

\textbf{Contributors}
\begin{tasks}[
            style=itemize,
            label-width=1em,
            column-sep=3.5em,
            before-skip=1.5ex,
            after-item-skip=1.5ex,
            ](2)
    \task Jiaxian Li
    \task Yuan Liang
    \task Yuqing Wang
    \task Qianben Chen
    \task Tiannan Wang
    \task Zhicong Lu
    \task Jiawei Ma
    \task Yuchen Eleanor Jiang
\end{tasks}


\textbf{Corresponding Author}
\begin{tasks}[
            style=itemize,
            label-width=1em,
            column-sep=3.5em,
            before-skip=1.5ex,
            after-item-skip=1.5ex 
            ](2)
    \task Wangchunshu Zhou
\end{tasks}

\vspace{1.5ex}
\textbf{Project Responsibilities}

\begin{itemize}

    \item\textit{Memory Framework Design \& Performance Optimization:} Piaohong Wang, Motong Tian, and Qianben Chen.  

    \item\textit{Paper Writing:} Piaohong Wang, Motong Tian, Tiannan Wang, Jiawei Ma, Zhicong Lu, and Qianben Chen.
    
    \item\textit{DeepResearch Benchmark Construction and Experiments:}  Yuan Liang, Jiaxian Li, Yuqing Wang, and Motong Tian. 

   \item\textit{Main Baseline Experiments:} Motong Tian, Jiaxian Li, and Yuan Liang. 
   
    \item\textit{Other Main Experiments and Results Analysis:} Motong Tian, and Piaohong Wang.    
    
\end{itemize}

\vfill 
\noindent\textsuperscript{*}These two authors contributed equally to this work. \\
\newpage
\section{Appendix}

\subsection{Evaluation Prompt}
\label{appendix:exp_prompts}
\begin{tcolorbox}[breakable,title=Prompt for Goal Alignment Criteria Generation]
You are an experienced research article evaluation expert. You excel at breaking down abstract evaluation dimensions (such as "Goal Understanding and Personalization Insight") into actionable, clear evaluation criteria tailored to the specific research task and user persona, and assigning reasonable weights with explanations for each criterion.\\
\texttt{\textless /system\_role\textgreater}

\vspace{1em}
\texttt{\textless user\_prompt\textgreater}\\
\textbf{Background}: We are evaluating a research article written for the following research task under the dimension of Goal Alignment.\\
\textbf{Goal Alignment:} Whether the research fully and accurately understands the relationship between the task and the user persona, extracts deep and implicit needs,
and generates a personalized report based on that understanding, with a focus on performing user-centered, deeply personalized matching between the user persona and task requirements.

\vspace{0.5em}
\texttt{\textless task\textgreater}\\
"\{task\_prompt\}"\\
\texttt{\textless /task\textgreater}
\vspace{0.5em}

The user persona is as follows:\\
\texttt{\textless persona\textgreater}\\
"\{persona\_prompt\}"\\
\texttt{\textless /persona\textgreater}

\vspace{0.5em}
\texttt{\textless instruction\textgreater}\\
Your goal:\\
For the Goal Alignment dimension of this research article, formulate a set of detailed, specific, and highly targeted evaluation criteria that are tightly aligned with the above \texttt{\textless task\textgreater} and \texttt{\textless persona\textgreater}. You need to:
\begin{enumerate}
    \item Deeply analyze the user persona and task scenario: Thoroughly examine the background characteristics, knowledge structure, cognitive habits, and latent expectations of \texttt{\textless persona\textgreater}. Combine this with the specific application scenario of \texttt{\textless task\textgreater} to identify the user’s core explicit needs and deeper implicit needs.
    \item Formulate personalized evaluation criteria: Based on the above analysis, propose specific evaluation criteria that reflect a deep understanding of \texttt{\textless persona\textgreater} and a close fit to the \texttt{\textless task\textgreater} scenario. These criteria should assess whether the content is well adapted to the user persona in style, depth, perspective, and practicality.
    \item Explain the personalization rationale: Provide a brief explanation (explanation) for each criterion, clarifying how it addresses the specific attributes of \texttt{\textless persona\textgreater} or special requirements of \texttt{\textless task\textgreater}, and why such targeting is critical to achieving a good match.
    \item Assign rational weights: Assign a weight (weight) to each criterion, ensuring that the total sum is 1.0. The distribution of weights should directly reflect the relative importance of each criterion in measuring how well the content matches "this particular user" in "this particular task." The closer a criterion is tied to persona characteristics and task scenario, the higher its weight should be.
\end{enumerate}

Core requirements:
\begin{enumerate}
    \item Deep personalization orientation: The analysis, criteria, explanations, and weights must be deeply rooted in the uniqueness of \texttt{\textless persona\textgreater} (e.g., their professional background, cognitive level, decision-making preferences, emotional needs) and the specific context of \texttt{\textless task\textgreater}. Avoid generic or templated evaluation.
    \item Focus on contextual responsiveness and resonance: The criteria should evaluate whether the content not only responds to the task at the informational level but also resonates with the context and expectations implied by the user persona in terms of expression style, reasoning logic, case selection, and level of detail.
    \item Rationale must reflect targeting: The \texttt{\textless analysis\textgreater} section must clearly explain how key features were extracted from the given \texttt{\textless persona\textgreater} and \texttt{\textless task\textgreater} to form these personalized criteria. Each criterion’s explanation must directly show how it serves this specific user and task.
    \item Weights must reflect personalization priorities: The weight distribution must logically demonstrate which aspects of alignment are the most critical success factors for "this user" completing "this task."
    \item Standard output format: Strictly follow the example format below. First output the \texttt{\textless analysis\textgreater} text, then immediately provide the \texttt{\textless json\_output\textgreater}.
\end{enumerate}
\texttt{\textless /instruction\textgreater}

\vspace{1em}
\texttt{\textless example\_rational\textgreater}\\
The example below demonstrates \textbf{how to develop Goal Alignment evaluation criteria based on the task requirements}. Focus on understanding the \textbf{thinking process and analytical approach} used in the example, rather than simply copying its content or numerical weights.\\
\texttt{\textless /example\_rational\textgreater}

...

\vspace{1em}
Please strictly follow the above instructions and methodology. Now, for the following specific task, start your work:\\
\texttt{\textless task\textgreater}\\
"\{task\_prompt\}"\\
\texttt{\textless /task\textgreater}

\vspace{0.5em}
\texttt{\textless persona\textgreater}\\
"\{persona\_prompt\}"\\
\texttt{\textless /persona\textgreater}

\vspace{0.5em}
Please output your \texttt{\textless analysis\textgreater} and \texttt{\textless json\_output\textgreater}.\\
\texttt{\textless /user\_prompt\textgreater}
\end{tcolorbox}

\begin{tcolorbox}[breakable,title=Prompt for Content Alignment Criteria Generation]
You are an experienced research article evaluation expert. You are skilled at breaking down abstract evaluation dimensions (such as ``Content Alignment'') into actionable, clear, and specific evaluation criteria tailored to the given research task and user persona, and assigning reasonable weights and explanations for each criterion.

\textless{}/system\_role\textgreater{}

\textless{}user\_prompt\textgreater{}

\textbf{Background}: We are providing a personalized scoring rubric for a specific task and user persona from the dimension of \textbf{Content Alignment}.

\textbf{Content Alignment}: Whether the research content is customized based on the user's interests, knowledge background, and other preferences.

\vspace{1em}

\textless{}task\textgreater{}

``\{task\_prompt\}''

\textless{}/task\textgreater{}

\vspace{1em}

The user persona is as follows:

\textless{}persona\textgreater{}

``\{persona\_prompt\}''

\textless{}/persona\textgreater{}

\textless{}instruction\textgreater{}

\textbf{Your Goal}: For the \textbf{Content Alignment} dimension of this research article, create a set of detailed, concrete, and highly tailored evaluation criteria for the above \textless{}task\textgreater{} and \textless{}persona\textgreater{}. You need to:
\begin{enumerate}
    \item \textbf{Analyze the Task and Persona}: Deeply analyze \textless{}task\textgreater{} and \textless{}persona\textgreater{} to infer the user's potential interests, knowledge background, and the depth and breadth of content they may prefer.
    \item \textbf{Formulate Criteria}: Based on your analysis, propose specific evaluation criteria that focus on whether the report's content matches the user's interest points and knowledge level.
    \item \textbf{Provide Explanations}: For each criterion, provide a brief explanation (\texttt{explanation}) explaining why it is important for evaluating the content alignment for this \textless{}task\textgreater{}.
    \item \textbf{Assign Weights}: Assign a reasonable weight to each criterion (\texttt{weight}), ensuring that the sum of all weights equals exactly 1.0. The weight allocation should logically reflect the personalization-first principle: criteria directly tied to unique personal traits, exclusive preferences, or specific contextual needs in the user persona should receive higher weights, as they are key to achieving true personalized content alignment.
    \item \textbf{Avoid Overlap}: Make sure the evaluation criteria focus solely on the \textbf{Content Alignment} dimension, avoiding overlap with other dimensions such as Goal Alignment, Expression Style Alignment, and Practicality/Actionability.
\end{enumerate}

\textbf{Core Requirements}:
\begin{enumerate}
    \item \textbf{Strongly Linked to the Persona}: The analysis, criteria, explanations, and weights must be directly connected to the user's interests, knowledge background, or content preferences.
    \item \textbf{Focus on Content Selection and Depth}: The criteria should assess whether the choice of content is precise and whether the depth is appropriate, rather than merely evaluating whether information is presented.
    \item \textbf{Provide Sufficient Rationale}: The \textless{}analysis\textgreater{} section must clearly articulate the overall reasoning behind formulating these criteria and weights, linking them to \textless{}task\textgreater{} and \textless{}persona\textgreater{}. Each \texttt{explanation} must clarify why the individual criterion is relevant.
    \item \textbf{Reasonable Weighting}: The weight distribution should be logical, reflecting the relative importance of each criterion in measuring content alignment, with particular emphasis on giving higher priority to personalized aspects.
    \item \textbf{Standardized Output Format}: Strictly follow the format below — output the \textless{}analysis\textgreater{} text first, immediately followed by \textless{}json\_output\textgreater{}.
\end{enumerate}
\textless{}/instruction\textgreater{}

\textless{}example\_rational\textgreater{}

The following example demonstrates \textbf{how to formulate content alignment evaluation criteria based on the task requirements and user persona}. Pay close attention to the \textbf{thinking process and analytical approach} in this example, rather than simply copying the content or weight values.

\textless{}/example\_rational\textgreater{}

…

Please strictly follow the above instructions and methodology. Now, for the following specific task, start your work:

\textless{}task\textgreater{}

``\{task\_prompt\}''

\textless{}/task\textgreater{}

\textless{}persona\textgreater{}

``\{persona\_prompt\}''

\textless{}/persona\textgreater{}

Please output your \textless{}analysis\textgreater{} and \textless{}json\_output\textgreater{}.

\textless{}/user\_prompt\textgreater{}
\end{tcolorbox}

\begin{tcolorbox}[breakable,title=Scoring Prompt for Personalization]
(For convenience and under time constraints, a temporary, unrefined prompt was employed for scoring during the experiment. The additional personalized indicators included in this temporary prompt—beyond the core metrics of goal alignment and content alignment—were ultimately discarded due to conceptual overlap. Therefore, this provisional prompt is functionally equivalent to our intended final evaluation design.)\\

\textless{}system\_role\textgreater{}You are a strict, meticulous, and objective expert in evaluating personalized research articles. You excel at deeply evaluating research articles based on specific personalization assessment criteria, providing precise scores and clear justifications.\textless{}/system\_role\textgreater{}

\bigskip
\textless{}user\_prompt\textgreater{}

\textbf{Task Background}\par
You are given an in-depth research task. Your job is to evaluate a research article written for this task in terms of its performance in \textbf{\textquotedbl{}Personalization Alignment\textquotedbl{}}. We will evaluate it across the following four dimensions:
\begin{enumerate}
    \item Goal Alignment
    \item Content Alignment
    \item Presentation Fit
    \item Actionability \& Practicality
\end{enumerate}

\textless{}task\textgreater{}\par
\textquotedbl{}\{task\_prompt\}\textquotedbl{}\par
\textless{}/task\textgreater{}

\bigskip
\textbf{User Persona}\par
\textless{}persona\textgreater{}\par
\textquotedbl{}\{persona\_prompt\}\textquotedbl{}\par
\textless{}/persona\textgreater{}

\bigskip
\textbf{Article to be Evaluated}\par
\textless{}target\_article\textgreater{}\par
\textquotedbl{}\{article\}\textquotedbl{}\par
\textless{}/target\_article\textgreater{}

\bigskip
\textbf{Evaluation Criteria}\par
You must evaluate the specific performance of this article in terms of personalization alignment, \textbf{following the criteria list below}, outputting your analysis and then assigning a score from 0--10. Each criterion includes its explanation, which you should read carefully.

\textless{}criteria\_list\textgreater{}\par
\{criteria\_list\}\par
\textless{}/criteria\_list\textgreater{}

\bigskip
\textless{}Instruction\textgreater{}\par
\textbf{Your Task}\par
Strictly follow \textbf{each criterion} in \textless{}criteria\_list\textgreater{} to evaluate how \textless{}target\_article\textgreater{} meets that criterion. You must:
\begin{enumerate}
    \item \textbf{Analyze Each Criterion}: For each item in the list, think about how the article meets the requirements of that criterion.
    \item \textbf{Analytical Evaluation}: Combine the article content, the task, and the user persona to analyze the article’s performance for that criterion, pointing out both strengths and weaknesses.
    \item \textbf{Scoring}: Based on your analysis, give a score between 0 and 10 (integer) for the article's performance on that criterion.
\end{enumerate}

\textbf{Scoring Rules}\par
For each criterion, give a score between 0 and 10 (integer). The score should reflect the quality of the article’s performance:
\begin{itemize}
    \item 0--2 points: Very poor. Almost completely fails to meet the requirement.
    \item 2--4 points: Poor. Meets the requirement only partially, with significant shortcomings.
    \item 4--6 points: Average. Basically meets the requirement; neither particularly good nor bad.
    \item 6--8 points: Good. Mostly meets the requirement, with notable strengths.
    \item 8--10 points: Excellent/Outstanding. Fully or exceptionally meets the requirement.
\end{itemize}

\textbf{Output Format Requirements}\par
Strictly follow the \textless{}output\_format\textgreater{} below to output the evaluation results for \textbf{each criterion}. \textbf{Do not include any irrelevant content, introductions, or conclusions}. Start from the first dimension and output all dimensions and their criteria in sequence:

\textless{}/Instruction\textgreater{}

\bigskip
\textless{}output\_format\textgreater{}
\begin{lstlisting}[]
{
    "goal_alignment": [
        {
            "criterion": "[The text of the first
            Goal Alignment criterion]",
            "analysis": "[Analysis]",
            "target_score": "[integer score 0-10]"
        },
        {
            "criterion": "[The text of the second 
            Goal Alignment criterion]",
            "analysis": "[Analysis]",
            "target_score": "[integer score 0-10]"
        },
        ...
    ],
    "content_alignment": [
        {
            "criterion": "[The text of the first Content Alignment 
            criterion]",
            "analysis": "[Analysis]",
            "target_score": "[integer score 0-10]"
        },
        ...
    ],
    "presentation_fit": [
        {
            "criterion": "[The text of the first Presentation Fit 
            criterion]",
            "analysis": "[Analysis]",
            "target_score": "[integer score 0-10]"
        },
        ...
    ],
    "actionability_practicality": [
        {
            "criterion": "[The text of the first Actionability
            & Practicality criterion]",
            "analysis": "[Analysis]",
            "target_score": "[integer score 0-10]"
        },
        ...
    ]
}
\end{lstlisting}
\textless{}/output\_format\textgreater{}
\end{tcolorbox}

\begin{tcolorbox}[breakable,title=Prompt for Persona Align Score]

\texttt{\textless system\_role\textgreater}\\
You are an experienced user research expert, skilled in analyzing and comparing user personas. Your task is to carefully compare a "Preset User Persona" and a "System Dynamically Learned User Persona", and identify the key similarities and differences between them.\\\texttt{\textless /system\_role\textgreater}

\vspace{1em}
\texttt{\textless user\_prompt\textgreater}\\
\textbf{Your analysis must strictly follow these four dimensions:}
\begin{enumerate}
\item**Basic Attributes \& Goals**: Compare similarities and differences in areas such as occupation, identity, core objectives, and usage motivations.
\item**Behavioral Patterns**: Compare similarities and differences in areas such as usage frequency, commonly used features, and interaction depth.
\item**Needs \& Preferences**: Compare similarities and differences in areas such as content preferences, feature requirements, and pain points.
\item**Overall Image Differences**: Summarize the overall perceptual differences between the two personas (e.g., "Diligent Learner" vs. "Efficient Problem Solver").
\end{enumerate}

\#\# Input Data:

- **Preset User Persona**:
  \{preset\_persona\_text\}

- **System Learned User Persona**:
  \{learned\_persona\_text\}

\#\# Output Requirements:

Please output your analysis results in **pure JSON format** only, without any additional explanations. The JSON structure should be as follows:
\begin{verbatim}
{{
  "comparison_by_dimension": {{
    "Basic Attributes & Goals": {{
      "Preset Persona Summary": "one-sentence summary",
      "Learned Persona Summary": "one-sentence summary",
      "Key Similarities": ["point 1", "point 2", ...],
      "Key Differences": ["point 1", "point 2", ...],
      "Difference Level": "High/Medium/Low" 
      // Judge based on the significance of differences
    }},
    "Behavioral Patterns": {{
      ... // Same structure as above
    }},
    "Needs & Preferences": {{
      ... // Same structure as above
    }}
  }},
  "overall_summary": {{
    "Preset Persona Overall Image": "a descriptive label or phrase",
    "Learned Persona Overall Image": "a descriptive label or phrase",
    "Overall Alignment Score": "integer from 0-100", // 100 
    indicates complete alignment
    "Most Important Insight": "one or two sentences explaining the 
    most critical insight"
  }}
}}
\end{verbatim}

\texttt{\textless /user\_prompt\textgreater}
\end{tcolorbox}

\subsection{Interaction Understanding Prompt}

\begin{tcolorbox}[breakable,title=UNDERSTAND USER EXPERIENCE PROMPT]
 Perform topic tagging on this message from user following these rules:

    1. Generate machine-readable tag\\
    2. Tag should cover:\\
       - Only one primary event concerning the user messages.\\
       - The author's attitude towards the event.\\
       - The topic should be the subject of the message which the user held attitude towards.\\
       - The topic and reason behind the attitude, sometimes you need to infer the attitude from the users' words. \\
       - The facts or events infered or revealed from the user's message.\\
       - If the author mention the time of the facts or events, the tag should also include the time inferred from the message (e.g., last day, last week)\\
       - Any attributes of the user revealed by the user's message (e.g., demographic features,biographical information,etc).\\
    3. Use this JSON format:
\begin{lstlisting}
{{
"text": "original message",
"tags": {{
"topic": ["event"],
"attitude": ["attitude towards the event":Postive or Negative or Mixed]
"reason" :["The reason concenring the attitude towards the event"]
"facts": ["The facts or events infered from the user's message"]
"attributes": ["The attributes of the user revealed by
the user's message"]
}},
"summary": "One sentence summary of the message"
"rationale": "brief explanation concenring why raising these tags"
}}
\end{lstlisting}
Example Input: "The jazz workshop helped me overcome performance anxiety"\\
Example Output:
\begin{lstlisting}
{{
"text": "Last week's jazz workshop helped me overcome
performance anxiety since the tutors are so patients.",
"tags": {{
    "topic": ["music workshop"],
    "attitude": ["Positive"],
    "reason": ["The tutors can teach the use patiently."],
    "facts":["join jazz workshop last week"],
    "attributes": ["user worrys about jazz performance"]
}},
"summary": "Jazz workshop helped the user overcome 
performance anxiety."
"rationale": "The user's performance anxiety was
alleviated with the help of Jazz Workshop.
Therefore , he is positive towards Jazz Workshop."
}}
\end{lstlisting}

Example Input: "I stop playing basketball for this semester due to too much stress."\\
Example Output:
\begin{lstlisting}
{{
    "text": "The user step away from playing baskerball 
    due to too much stress.",
    "tags": {{
        "topic": ["playing basketball"],
        "attitude": ["negative"],
        "reason": ["Too much stree for playing basketball"],
        "facts:["stop playing basketball"],
        "attributes": ["user hate stress"]
    }},
    "summary": "The user stop playing baskerball due to 
    too much stress."
    "rationale": "The user stop playing baskerball due 
    to too much stress. 
    Therefore, the user is negative towards playing basketball."
}}
\end{lstlisting}

Example Input: "I go back to play basketball due to strenghten my body yesterday."\\
Example Output:
\begin{lstlisting}
{{
"text": "The user return to play basketball due to 
strenghten the body.",
"tags": {{
    "topic": ["playing basketball"],
    "attitude": ["Positive"],
    "reason": ["Baskterball could help strenghtening the body"],
    "facts":["return to play basketball yesterday"],
    "attributes": ["User value the body"]
}},
"summary": "The user go back to play basketball due to 
strenghten the body."
"rationale": "he user go back to play basketball due to
strenghten the body. There, the user is positive towards 
playing basketball."
}}
\end{lstlisting}
Example Input: "I hate playing basktetball due to its preasure"\\
Example Output:
\begin{lstlisting}
{{
"text": "I hate playing basktetball since I move from my
hometowm GuangZhou due to its preasure.",
"tags": {{
    "topic": ["hate playing basketball"],
    "attitude": ["negative"],
    "reason": ["The user hates playing basktetball for preasure."],
    "facts":["hate playing basketball"],
    "attributes": ["user hate stress","user's hometown 
    is GuangZhou"]
}},
"summary": "The user go back to play basketball due to
strenghten the body."
"rationale": "The user go back to play basketball due to
strenghten the body. There, the user is positive towards 
playing basketball."
}}
\end{lstlisting}
Now analyze this message:
"{message}"

\end{tcolorbox}

\newpage

\subsection{\method{} Workflow Visualization}

\begin{figure}[h]
    \centering
\includegraphics[width=1\textwidth]{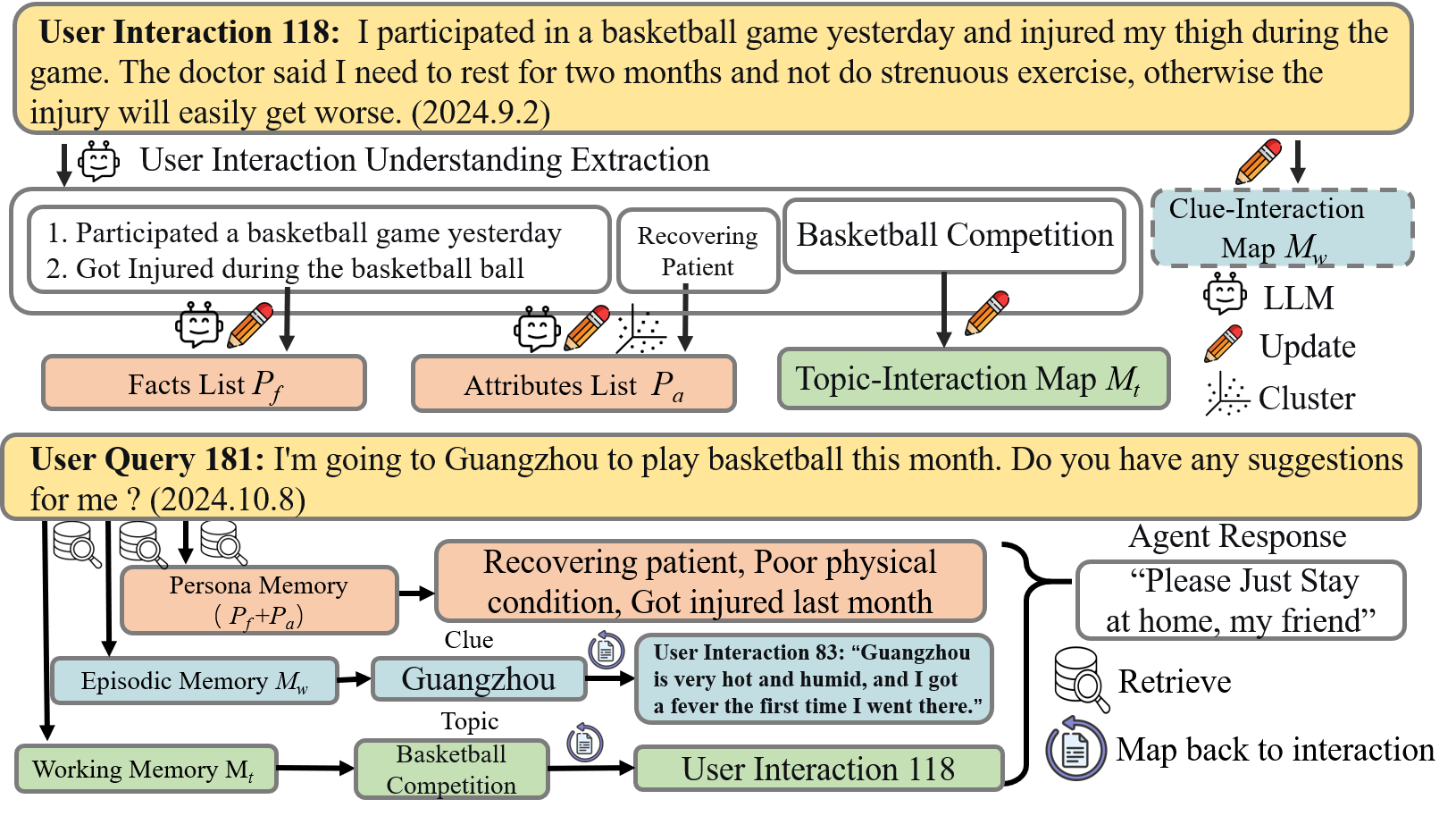}
    \caption{Top: The process of encoding user interactions into memory in \method{}. Different colors refers to different memory components. \method{} encodes a user interaction into memory by extracting and recording relevant user attributes and event data into {\color{atomictangerine}{persona memory}},{\color{babyblue}{episodic memory}}, and {\color{asparagus}{working memory}}. Bottom: The memory retrieval process concerning one user interaction in \method{}. \method{} retrieves from all its three memory components concerning one new user query.}
    \label{method}
\end{figure}

\newpage
\appendix
\addcontentsline{toc}{section}{Appendix}
\addtocontents{toc}{\protect\setcounter{tocdepth}{0}}

\end{document}